\documentclass[11pt,a4paper]{article}
\usepackage[hyperref]{acl2019}
\usepackage{times}
\usepackage{latexsym}

\usepackage{url}

\usepackage{color}
\usepackage{tabularx}
\usepackage{amsmath}
\usepackage{amssymb}
\usepackage{graphicx}
\usepackage{multirow}
\usepackage{here}
\usepackage{setspace}
\usepackage{tikz,tikz-qtree}
\usetikzlibrary{positioning,arrows,shapes}
\usepackage{algorithm}
\usepackage{algorithmic}
\usepackage{proof}
\usepackage{gb4ea}
\usepackage{ebproof}
\usepackage{textcomp}
\usepackage[T1]{fontenc}

\usepackage{url}
\usepackage{mediabb}
\aclfinalcopy 

\newcommand{\reds}{red!80!black}
\newcommand{\blues}{blue!80!black}

\newcommand{\np}[1]{[$_{\scriptsize\mathsf{NP}}$\,#1]}
\newcommand{\vp}[1]{[$_{\scriptsize\mathsf{VP}}$\,#1]}

\newcommand{\positive}[1]{#1\,$\textcolor{\reds}{\uparrow}$}
\newcommand{\positiveI}[1]{#1$\textcolor{\reds}{\uparrow}$}
\newcommand{\negativeI}[1]{#1$\textcolor{\blues}{\downarrow}$}
\newcommand{\negative}[1]{#1\,$\textcolor{\blues}{\downarrow}$}

\tikzset{
var node/.style={circle,draw,node distance=2.0cm}, 
exvar node/.style={dotted,circle,draw,node distance=2.0cm,thick}, 
pred node/.style={rectangle,rounded corners,draw,node distance=1.7cm}, 
rel node/.style={fill=white,circle,inner sep=0.1mm}
}

\newcommand{\final}[1]{\textcolor{black}{#1}}
\newcommand{\modelcite}[1]{\citeauthor{#1},\ \citeyear{#1}}

\title{Can neural networks understand monotonicity reasoning?}

\author{
  \parbox{\linewidth}{\centering
   Hitomi Yanaka$^{1,2}$,
   Koji Mineshima$^2$,
   Daisuke Bekki$^2$,\linebreak
   Kentaro Inui$^{1,3}$,
   Satoshi Sekine$^1$,
   Lasha Abzianidze$^4$, and
   Johan Bos$^4$
  }
  \\
   $^1$\mbox{\rm RIKEN,}
   $^2$\mbox{\rm Ochanomizu University,}
   $^3$\mbox{\rm Tohoku University, Japan} \\ 
$^4$\mbox{\rm University of Groningen, Netherlands}
  \\
  \parbox{\linewidth}{\centering
   {\tt \{hitomi.yanaka, satoshi.sekine\}@riken.jp},
   {\tt mineshima.koji@ocha.ac.jp},
   {\tt bekki@is.ocha.ac.jp},
   {\tt inui@ecei.tohoku.ac.jp},
   {\tt \{l.abzianidze, johan.bos\}@rug.nl}
   }
}

\date{}
\begin{document}
\maketitle
\begin{abstract}
Monotonicity reasoning is one of the important reasoning skills for any intelligent natural language inference (NLI) model in that it requires the ability to capture the interaction between lexical and syntactic structures.
Since no test set has been developed for monotonicity reasoning with wide coverage, it is still unclear whether neural models can perform monotonicity reasoning in a proper way.
To investigate this issue, we introduce the Monotonicity Entailment Dataset (MED).
Performance by state-of-the-art NLI models on the new test set is substantially worse, under 55\%, especially on downward reasoning.
In addition, analysis using a monotonicity-driven data augmentation method showed that these models might be limited in their generalization ability in upward and downward reasoning.

\end{abstract}
\section{Introduction}
\label{sec:intro}
Natural language inference (NLI), also known as recognizing textual entailment (RTE), has been proposed as a benchmark task for natural language understanding.
Given a premise $P$
and a hypothesis $H$,
the task is to determine whether the premise semantically entails the hypothesis~\cite{series/synthesis/2013Dagan}.
A number of recent works attempt to test and analyze what type of inferences an NLI model may be performing,
focusing on various types of lexical inferences~\cite{glockner-shwartz-goldberg:2018:Short,naik-EtAl:2018:C18-1,poliak-EtAl:2018:S18-2}
and logical inferences~\cite{Bowman2015,Evans2018CanNN}.

Concerning logical inferences, monotonicity reasoning~\cite{10.2307/25001141,moss2014}, which is a type of reasoning based on word replacement, requires the ability to capture the interaction between lexical and syntactic structures.
Consider examples in (\ref{ex:1}) and (\ref{ex:2}).

{\footnotesize
\begin{exe}
\ex \label{ex:1}
\begin{xlist}
\ex \label{ex:1a} \textit{\textbf{All}} [\,\negative{\textit{\underline{workers}}}] [\positive{\textit{joined for a \underline{French dinner}}}]
\ex \label{ex:1b} \textit{All workers joined for a \underline{dinner}}
\ex \label{ex:1c} \textit{All \underline{new workers} joined for a French dinner}
\end{xlist}
\ex \label{ex:2}
\begin{xlist}
\ex \label{ex:2a} \textit{\textbf{Not all}} [\positive{\textit{\underline{new workers}}}] \textit{joined for a dinner}
\ex \label{ex:2b} \textit{Not all \underline{workers} joined for a dinner}
\end{xlist}
\end{exe}
}

\noindent 
A context is \textbf{upward entailing}
(shown by [...\positive{}])
that allows
an inference from (\ref{ex:1a}) to (\ref{ex:1b}),
where \textit{French dinner}
is replaced by a more general concept \textit{dinner}.
On the other hand, a \textbf{downward entailing} 
context (shown by [...\negative{}])
allows
an inference from (\ref{ex:1a}) to (\ref{ex:1c}),
where \textit{workers} is replaced by
a more specific concept \textit{new workers}.
Interestingly,
the direction of monotonicity can be reversed again by embedding yet another downward entailing context (e.g., \textit{not} in (\ref{ex:2})), as witness the fact that (\ref{ex:2a}) entails (\ref{ex:2b}). 
To properly handle both directions of monotonicity, NLI models must detect monotonicity operators
(e.g., \textit{all}, \textit{not})
and their arguments from the syntactic structure.

For previous datasets containing monotonicity inference problems, FraCaS~\cite{cooper1994fracas} and the GLUE diagnostic dataset~\cite{wang2018glue} are 
manually-curated
datasets for testing a wide range of linguistic phenomena.
However, monotonicity problems are limited to very small sizes (FraCaS: 37/346 examples and GLUE: 93/1650 examples).
The limited syntactic patterns and vocabularies in previous test sets are obstacles in accurately evaluating NLI models on monotonicity reasoning.

To tackle this issue, we present a new evaluation dataset\footnote{The dataset will be made publicly available at https://github.com/verypluming/MED.} that covers a wide range of monotonicity reasoning that was created by crowdsourcing and collected from linguistics publications (Section~\ref{sec:data}).
Compared with manual or automatic construction, we can collect naturally-occurring examples by crowdsourcing and well-designed ones from linguistics publications.
To enable the evaluation of skills required for monotonicity reasoning, we annotate each example in our dataset with linguistic tags associated with monotonicity reasoning.

We measure the performance of state-of-the-art NLI models on monotonicity reasoning and investigate their generalization ability in upward and downward reasoning (Section~\ref{sec:result}).
The results show that all models trained with
SNLI~\cite{Bowman2015} and MultiNLI~\cite{DBLP:journals/corr/WilliamsNB17}
perform worse on downward inferences than on upward inferences.

\final{In addition, we analyzed the performance of models trained  with an automatically created monotonicity dataset, HELP~\cite{yanaka2019}.}
The analysis with monotonicity data augmentation shows that models tend to perform better in the same direction of monotonicity with the training set, while they perform worse in the opposite direction.
This indicates that the accuracy on monotonicity reasoning depends solely on the majority direction in the training set, and models might lack the ability to capture the structural relations between monotonicity operators and their arguments.

\section{Monotonicity}
\label{sec:background}

As an example of a monotonicity inference,
consider the example with the determiner \textit{every} in (\ref{ex:3});
here the premise $P$ entails the hypothesis $H$.

{\footnotesize
\begin{exe}
\ex \label{ex:3}
 \begin{xlist}
    \exi{$P$:} \label{ex:3a} \textit{Every} \np{\negative{\textit{\underline{person}}}} \vp{\positive{\textit{bought a \underline{movie ticket}}}}
    \exi{$H$:} \label{ex:3b} \textit{Every \underline{young person} bought a \underline{ticket}}
 \end{xlist}
\end{exe}
}

\noindent \textit{Every} is downward entailing in the first argument ($\mathsf{NP}$) and upward entailing in the second argument ($\mathsf{VP}$),
and thus the term \textit{person} can be more specific by adding modifiers (\textit{person} $\sqsupseteq$ \textit{young person}),
replacing it with its hyponym (\textit{person} $\sqsupseteq$ \textit{spectator}),
or adding conjunction (\textit{person} $\sqsupseteq$ \textit{person and alien}).
On the other hand, the term \textit{buy a ticket} can be more general by removing modifiers (\textit{bought a movie ticket} $\sqsubseteq$ \textit{bought a ticket}), replacing it with its hypernym (\textit{bought a movie ticket} $\sqsubseteq$ \textit{bought a show ticket}), or adding disjunction (\textit{bought a movie ticket} $\sqsubseteq$ \textit{bought or sold a movie ticket}).
Table~\ref{tab:downex} shows
determiners modeled as binary operators
and their polarities with respect to
the first and second arguments.

\begin{table}[bt]
\scalebox{0.75}{
  \begin{tabular}{lll} \hline
    Determiners&First argument&Second argument\\ \hline \hline
    \textit{every, each, all}&downward&upward\\ \hline
    \textit{some, a, a few, many,}&\multirow{2}{*}{upward}&\multirow{2}{*}{upward}\\ 
    \textit{several}, proper noun&&\\ \hline
    \textit{any, no, few, at most X,}&\multirow{2}{*}{downward}&\multirow{2}{*}{downward}\\ 
    \textit{fewer than X, less than X}&&\\ \hline
    \textit{the, both, most, this, that}&non-monotone&upward\\ \hline
    \textit{exactly}&non-monotone&non-monotone\\ \hline

  \end{tabular}
}
\caption{\label{tab:downex} Determiners and their polarities.}
\end{table}

\begin{table}[bt]
\scalebox{0.81}{
  \begin{tabular}{ll} \hline
  Category&Examples\\ \hline \hline
    determiners & \textit{every, all, any, few, no}\\ \hline
    negation & \textit{not, n't, never} \\ \hline
    verbs & \textit{deny, prohibit, avoid} \\ \hline
    nouns & \textit{absence of, lack of, prohibition} \\ \hline
    adverbs & \textit{scarcely, hardly, rarely, seldom} \\ \hline
    prepositions & \textit{without, except, but} \\ \hline
    conditionals & \textit{if, when, in case that, provided that, unless}\\ \hline
  \end{tabular}
}
\caption{\label{tab:downex2} Examples of downward operators.}
\end{table}





There are various types of downward operators, not limited to determiners (see Table~\ref{tab:downex2}).
As shown in (\ref{ex:4}), if a propositional object is embedded in a downward monotonic context (e.g., \textit{when}), the polarity of words over its scope can be reversed.

{\footnotesize
\begin{exe}
\ex \label{ex:4}
 \begin{xlist}
    \exi{$P$:} 
\textit{When} [\textit{every} \np{\positive{\textit{young person}}} \vp{\negative{\textit{bought a ticket}}}], [\textit{that shop was open}]
\exi{$H$:} \textit{When} [\textit{every} \np{\textit{person}} \vp{\textit{bought a movie ticket}}], [\textit{that shop was open}]
 \end{xlist}
\end{exe}
}
\noindent Thus, the polarity (\positiveI{} and \negativeI{}), where the replacement with more general (specific) phrases licenses entailment, needs to be determined by the interaction of monotonicity properties and syntactic structures;
polarity of each constituent is calculated based on a monotonicity operator
of functional expressions (e.g., \textit{every}, \textit{when}) and their function-term relations.

\section{Dataset}
\label{sec:data}
\subsection{Human-oriented dataset}
To create monotonicity inference problems, we should satisfy three requirements:
 (a) detect the monotonicity operators and their arguments;
 (b) based on the syntactic structure, induce the polarity of the argument positions; and
 (c) replace the phrase in the argument position with a more general or specific phrase in natural and various ways (e.g., by using lexical knowledge or logical connectives).
For (a) and (b), we first conduct polarity computation on a syntactic structure for each sentence, and then select premises involving upward/downward expressions. 

For (c), we use crowdsourcing to narrow or broaden the arguments.
The motivation for using crowdsourcing is to collect naturally alike monotonicity inference problems that include various expressions.
One problem here is that
it is unclear how to
\final{instruct workers to}
create monotonicity inference problems without knowledge of natural language syntax and semantics.
We must make tasks simple for workers to comprehend and provide sound judgements.
Moreover, recent studies~\cite{gururangan-EtAl:2018:N18-2,poliak-EtAl:2018:S18-2,DBLP:conf/lrec/Tsuchiya18} point out that previous crowdsourced datasets, such as SNLI~\cite{snli:emnlp2015} and MultiNLI~\cite{DBLP:journals/corr/WilliamsNB17}, include hidden biases.
As these previous datasets are motivated by approximated entailments, workers are asked to freely write hypotheses given a premise,
which does not strictly restrict them to creating logically complex inferences.

\begin{figure}
  \includegraphics[width=8cm, bb=0 0 497 320]{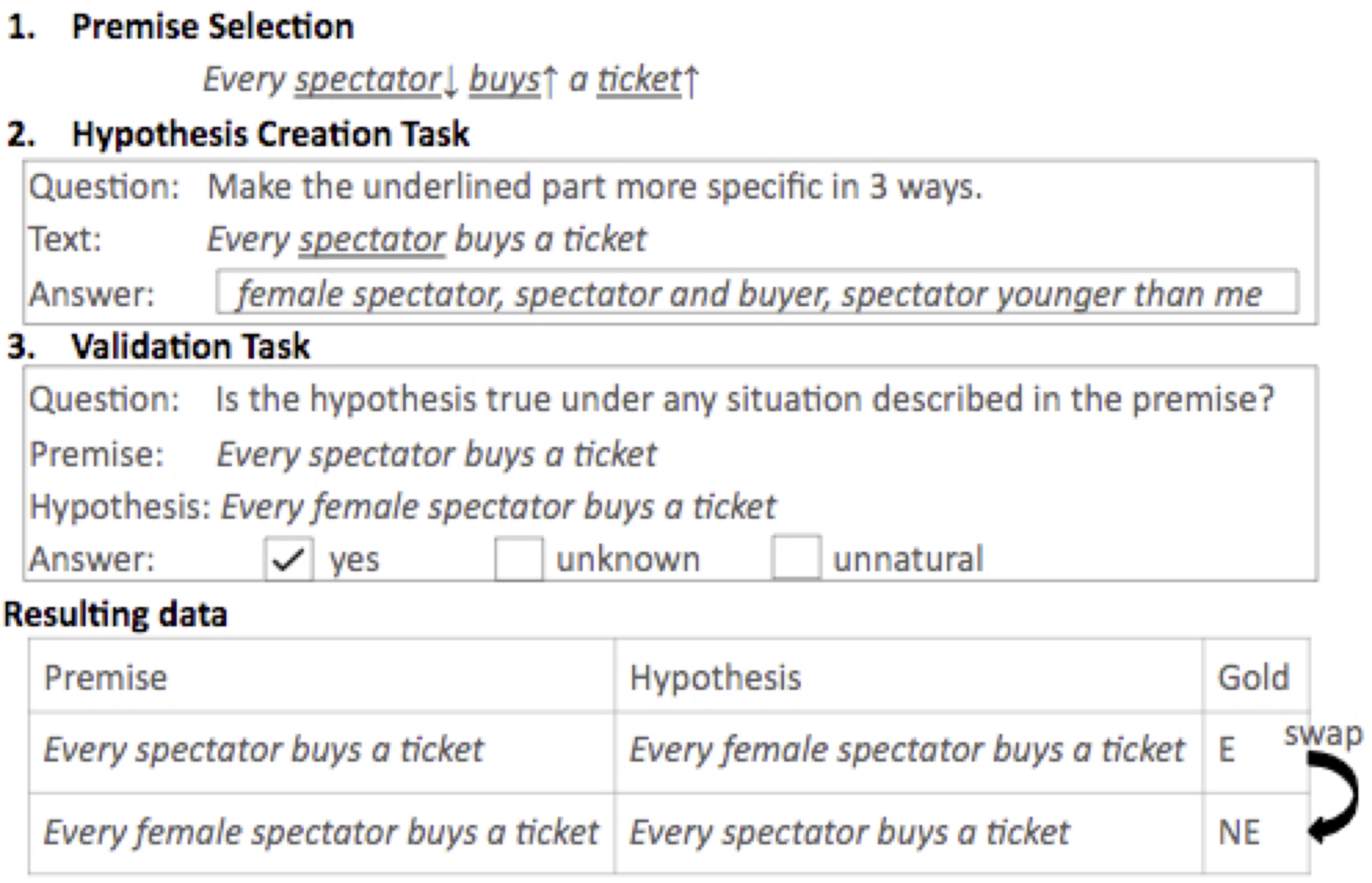}
  \caption{\label{tab:fig2}\final{Overview of our human-oriented dataset creation. E: entailment, NE: non-entailment.}}
\end{figure}

Taking these concerns into consideration, we designed two-step tasks to be performed via crowdsourcing for creating a monotonicity test set; (i) a hypothesis creation task and (ii) a validation task.
The task (i) is to create a hypothesis by making some polarized part of an original sentence more specific.
Instead of writing a complete sentence from scratch, workers are asked to rewrite only a relatively short sentence.
By restricting workers to rewrite only a polarized part, we can effectively collect monotonicity inference examples.
The task (ii) is to annotate an entailment label for the premise-hypothesis pair generated in (i).
\final{Figure~\ref{tab:fig2} summarizes the overview of our human-oriented dataset creation.}
We used the crowdsourcing platform Figure Eight for both tasks.

\subsubsection{Premise collection}
As a resource, we use declarative sentences with more than five tokens from the Parallel Meaning Bank (PMB)~\cite{abzianidze-EtAl:2017:EACLshort}.
The PMB contains syntactically correct sentences annotated with its syntactic category in Combinatory Categorial Grammar (CCG;~\modelcite{Steedman00}) format, which is suitable for our purpose. 
To get a whole CCG derivation tree, we parse each sentence by the state-of-the-art CCG parser,
depccg~\cite{yoshikawa-noji-matsumoto:2017:Long}.
Then, we add a polarity to every constituent of the CCG tree by the polarity computation system ccg2mono~\cite{humoss2018} and make the polarized part a blank field.

We ran a trial rephrasing task on 500 examples and detected 17 expressions that were too general and thus difficult to rephrase them in a natural way (e.g., \textit{every one}, \textit{no time}).
We removed examples involving such expressions.
To collect more downward inference examples, we select examples involving determiners in Table~\ref{tab:downex} and downward operators in Table~\ref{tab:downex2}.
As a result, we selected 1,485 examples involving expressions having arguments with upward monotonicity
and 1,982 examples involving expressions having arguments with downward monotonicity.


\subsubsection{Hypothesis creation}
We present crowdworkers with a sentence whose polarized part is underlined, and ask them to replace the underlined part with more specific phrases in three different ways.
In the instructions, we showed examples rephrased in various ways: by adding modifiers, by adding conjunction
phrases, and by replacing a word with its hyponyms.

Workers were paid US\$0.05 for each set of substitutions, and each set was assigned to three workers.
To remove low-quality examples, we set the minimum time it should take to complete each set to 200 seconds.
The entry in our task was restricted to workers from native speaking English countries.
128 workers contributed to the task, and we created 15,339 hypotheses (7,179 upward examples and 8,160 downward examples).

\subsubsection{Validation}
The gold label of each premise-hypothesis pair created in the previous task is automatically determined by monotonicity calculus.
That is, a downward inference pair is labeled as \textit{entailment}, while an upward inference pair is labeled as \textit{non-entailment}.

However, workers sometimes provided some ungrammatical or unnatural sentences such as the case \final{where a rephrased phrase does not satisfy the selectional restrictions (e.g., original: \textit{Tom doesn't live in \underline{Boston}}, rephrased: \textit{Tom doesn't live in \underline{yes}})},
making it difficult to judge their entailment relations.
Thus, we performed an annotation task to ensure accurate labeling of gold labels.
We asked workers about the entailment relation of each premise-hypothesis pair as well as how natural it is.

Worker comprehension of an entailment relation directly affects the quality of inference problems.
To avoid worker misunderstandings, we showed workers the following definitions of labels and five examples for each label:
\begin{enumerate}
  \item \textit{entailment}: the case where the \final{hypothesis} is true under any situation that the premise describes.
  \item \textit{non-entailment}: the case where the \final{hypothesis} is not always true under a situation that the premise describes.
  \item \textit{unnatural}: the case where either the premise and/or the \final{hypothesis} is ungrammatical or does not make sense.
\end{enumerate}

Workers were paid US\$0.04 for each question, and each question was assigned to three workers.
To collect high-quality annotation results, we imposed ten test questions on each worker, and removed workers who gave more than three wrong answers.
We also set the minimum time it should take to complete each question to 200 seconds.
1,237 workers contributed to this task, and we annotated gold labels of 15,339 premise-hypothesis pairs.

Table~\ref{tab:result2} shows the numbers of cases where answers matched gold labels automatically determined by monotonicity calculus.
This table shows that there exist inference pairs whose labels are difficult even for humans to determine; there are 3,354 premise-hypothesis pairs whose gold labels as annotated by polarity computations match with those answered by all workers.
We selected these \final{naturalistic monotonicity inference} pairs for the candidates of the final test set.

To \final{make the distribution of gold labels symmetric,}
we checked these pairs to determine if we can swap the premise and the hypothesis, reverse their gold labels, and create another monotonicity inference pair.
In some cases, shown below, the gold label cannot be reversed if we swap the premise and the hypothesis.

\begin{table}
\centering
\scalebox{0.63}{
\begin{tabular}{lrrr} \hline
        & Upward /cases(\%)& Downward /cases(\%)&Total /cases(\%)\\ \hline \hline
3 labels match & 1,069 (7.0)& 2,285 (14.9) & 3,354 (21.9)\\ 
2 labels match& 1,814 (11.8)& 2,301 (15.0) & 4,115 (26.8)\\ 
1 labels match& 2,295 (15.0)& 1,915 (12.5) & 4,210 (27.5)\\ 
no match& 1,998 (27.8)& 1,652 (10.8) & 3,650 (37.8)\\ \hline
\end{tabular}
}
\caption{\label{tab:result2} Numbers of cases where answers matched automatically determined gold labels.}
\vspace{-1.0em}
\end{table}

\begin{table*}
\begin{center}
\scalebox{0.72}{
\begin{tabular}{lllll} \hline
                          Genre&Tags& Premise & Hypothesis&Gold \\ \hline \hline
   \multirow{10}{*}{Crowd}&up&  \textit{There is a cat on the chair}&\textit{There is a cat sleeping on the chair}&NE \\ \cline{2-5}
   &up:& \textit{If you heard her speak English, you would take her} & \textit{If you heard her speak English, you would take her}& \\ 
   &cond& \textit{for a native American} &\textit{for an American}&E   \\ \cline{2-5}
   &up:rev:&\textit{Dogs and cats have all the good qualities of people} & \textit{Dogs have all the good qualities of people without}& \\ 
   &conj& \textit{without at the same time possessing their weaknesses} & \textit{at the same time possessing their weaknesses}&E   \\ \cline{2-5}
   &up:lex&\textit{He approached the boy reading a magazine}&\textit{He approached the boy reading a book}&E\\ \cline{2-5} 
   &down:lex& \textit{Tom hardly ever listens to music} & \textit{Tom hardly ever listens to rock 'n' roll}&E \\ \cline{2-5}
   &down:conj& \textit{You don't like love stories and sad endings} & \textit{You don't like love stories}&NE \\ \cline{2-5}
   &down:cond& \textit{If it is fine tomorrow, we'll go on a picnic} & \textit{If it is fine tomorrow in the field, we'll go on a picnic}&E\\ \cline{2-5}
   &down& \textit{I never had a girlfriend before} & \textit{I never had a girlfriend taller than me before}&E \\ \hline
   \multirow{7}{*}{Paper}
   &up:rev&  \textit{Every cook who is not a tall man ran}&\textit{Every cook who is not a man ran}&E  \\ \cline{2-5}
   &up:disj& \textit{Every man sang} &\textit{Every man sang or danced}&E   \\ \cline{2-5}
   &up:lex:& \textit{None of the sopranos sang with fewer than three of} &\textit{None of the sopranos sang with fewer than three of}&   \\
   &rev& \textit{the tenors} &\textit{the male singers}&E   \\ \cline{2-5} 
   &non&\textit{Exactly one man ran quickly} &\textit{Exactly one man ran}&NE \\ \cline{2-5}
   &down&\textit{At most three elephants are blue}&\textit{At most three elephants are navy blue}&E\\ \hline

\end{tabular}
}
\caption{\label{tab:ex} Examples in the MED dataset. Crowd: problems collected through crowdsourcing, Paper: problems collected from linguistics publications, up: upward monotone, down: downward monotone, non: non-monotone, cond: condisionals, rev: reverse, conj: conjunction, disj: disjunction, lex: lexical knowledge, E: entailment, NE: non-entailment.}
\end{center}
\end{table*}


\begin{table}
\centering
\scalebox{0.72}{
\begin{tabular}{llrrr} \hline
Type&Label&Crowd&Paper&Total\\ \hline \hline
\multirow{2}{*}{Upward (1,820)}&Entailment&323&305&628\\
&Non-entailment&893&299&1,192\\ \hline
\multirow{2}{*}{Downward (3,270)}&Entailment&1,871&201&2,072\\ 
&Non-entailment&979&219&1,198\\ \hline
\multirow{2}{*}{Non-monotone (292)}&Entailment&0&15&15\\
&Non-entailment&2&275&277\\ \hline
Total&&4,068&1,314&5,382\\ \hline
\end{tabular}
}
\caption{\label{tab:ling} \final{Statistics for the MED dataset.}}
\vspace{-1.0em}
\end{table}






\paragraph{(a) Replacement with synonyms}
In (\ref{ex:7}), \textit{child} and \textit{kid} are not hyponyms but synonyms, and the premise $P$ and the hypothesis $H$ are paraphrases.

{\footnotesize
\begin{exe}
\ex \label{ex:7}
 \begin{xlist}
    \exi{$P$:} \label{ex:7a} \textit{Tom is no longer a child}
    \exi{$H$:} \label{ex:7b} \textit{Tom is no longer a kid}
 \end{xlist}
\end{exe}
}

\noindent These cases are not strict downward inference problems,
in the sense that a phrase is not replaced by its
hyponym/hypernym.

\paragraph{(b) Non-constituents}
Consider the example (\ref{ex:8}).

{\footnotesize
\begin{exe}
\ex \label{ex:8}
 \begin{xlist}
    \exi{$P$:} \label{ex:8a} \textit{The moon has no atmosphere}
    \exi{$H$:} \label{ex:8b} \textit{The moon has no atmosphere, and the gravity force is too low}
 \end{xlist}
\end{exe}
}

\noindent The hypothesis $H$ was created by asking workers to make \textit{atmosphere} in the premise $P$ more specific.
However, the additional phrase \textit{and the gravity force is too low} does not form constituents with \textit{atmosphere}.
Thus, such examples
are not strict downward monotone inferences.

In such cases as (a) and (b), we do not swap the premise and the hypothesis.
In the end, we collected 4,068 examples from crowdsourced datasets.

\subsection{Linguistics-oriented dataset}
We also collect monotonicity inference problems from previous manually curated datasets and linguistics publications.
The motivation is that previous linguistics publications related to monotonicity reasoning are expected to contain well-designed inference problems, which might be challenging problems for NLI models.

We collected 1,184 examples from 11 linguistics publications~\cite{BarwiseCooper81,Hoeksema,HeimKratzer98,Asher,fyodorov03,Geurts03,GeurtsSlik,Zamansky06,Szabolcsi05,Winter,denic2019}.
Regarding previous manually-curated datasets, we collected 93 examples for monotonicity reasoning from the GLUE diagnostic dataset, and 37 single-premise problems from FraCaS.

Both the GLUE diagnostic dataset and FraCaS categorize problems by their types of monotonicity reasoning, but we found that each dataset has different classification criteria.\footnote{FraCaS categorizes each problem by whether its replacement broadens an argument (upward monotone) or narrows it (downward monotone).}
Thus, following GLUE, we reclassified problems into three types of monotone reasoning (upward, downward, and non-monotone) by checking if they include (i) the target monotonicity operator in both the premise and the hypothesis and (ii) the phrase replacement in its argument position.
In the GLUE diagnostic dataset, there are several problems whose gold labels are \textit{contradiction}.
We regard them as \textit{non-entailment} in that the premise does not semantically entail the hypothesis.

\subsection{Statistics}
We merged the human-oriented dataset created via crowdsourcing and the linguistics-oriented dataset created from linguistics publications to create the current version of the monotonicity entailment dataset (MED).
Table~\ref{tab:ex} shows some examples from the MED dataset.
We can see that our dataset contains various phrase replacements (e.g., conjunction, relative clauses, and comparatives).
Table~\ref{tab:ling} reports the statistics of the MED dataset, including 5,382 premise-hypothesis pairs (1,820 upward examples, 3,270 downward examples, and 292 non-monotone examples).
Regarding non-monotone problems, gold labels are always \textit{non-entailment}, whether a hypothesis is more specific or general than its premise, and thus almost all non-monotone problems are labeled as \textit{non-entailment}.\footnote{15 non-monotone problems which include the replacement with synonyms are labeled as \textit{entailment}.}
The size of the word vocabulary in the MED dataset is 4,023, and overlap ratios of vocabulary with previous standard NLI datasets is 95\% with MultiNLI and 90\% with SNLI.

We assigned a set of annotation tags for linguistic phenomena to each example in the test set.
These tags allow us to analyze how well models perform on each linguistic phenomenon related to monotonicity reasoning.
We defined 6 tags \final{(see Table~\ref{tab:ex} for examples):}
\begin{enumerate}
  \item \textit{lexical knowledge} (2,073 examples): inference problems that require lexical relations (i.e., hypernyms, hyponyms, or synonyms)
  \item \textit{reverse} (240 examples): inference problems where a propositional object is embedded in a downward environment more than once 
  \item \textit{conjunction} (283 examples): inference problems that include the phrase replacement by adding conjunction (\textit{and}) to the hypothesis
  \item \textit{disjunction} (254 examples): inference problems that include the phrase replacement by adding disjunction (\textit{or}) to the hypothesis 
  \item \textit{conditionals} (149 examples): inference problems that include conditionals (e.g., \textit{if, when, unless}) in the hypothesis
  \footnote{\final{\textit{When}-clauses can have temporal and non-temporal interpretations~\cite{steedman1988}.
  We assign the conditional tag to those cases where \textit{when} is interchangeable with \textit{if}, thus excluding those cases where \textit{when}-clauses have temporal episodic interpretation (e.g., \textit{When she came back from the trip, she bought a gift}).}}
  \item \textit{negative polarity items} (NPIs) (338 examples): inference problems that include NPIs (e.g., \textit{any, ever, at all, anything, anyone, anymore, anyhow, anywhere}) in the hypothesis
\end{enumerate}

\section{Results and Discussion}
\label{sec:result}
\subsection{Baselines}
To test the difficulty of our dataset,
we checked the majority class label and the accuracies of five state-of-the-art NLI models adopting different approaches:
BiMPM (Bilateral Multi-Perspective Matching Model;~\modelcite{Wang:2017:BMM:3171837.3171865}),
ESIM (Enhanced Sequential Inference Model;~\modelcite{chen-EtAl:2017:Long3}),
Decomposable Attention Model~\cite{D16-1244},
KIM (Knowledge-based Inference Model;~\modelcite{P18-1224}),
and BERT (Bidirectional Encoder Representations from Transformers model;~\modelcite{BERT2018new}).
Regarding BERT, we checked the performance of a model pretrained on Wikipedia and BookCorpus for language modeling and trained with SNLI and MultiNLI.
For other models, we checked the performance trained with SNLI.
In agreement with our dataset, we regarded the prediction label \textit{contradiction} as \textit{non-entailment}.

Table~\ref{tab:perform} shows that the accuracies of all models were better on upward inferences, in accordance with the reported results of the GLUE leaderboard.
The overall accuracy of each model was low.
In particular, all models underperformed the majority baseline on downward inferences, despite some models having rich lexical knowledge from a knowledge base (KIM) or pretraining (BERT).
This indicates that
downward inferences are difficult to perform even with the expansion of lexical knowledge.
In addition, it is interesting to see that if a model performed better on upward inferences, it performed worse on downward inferences.
We will investigate these results in detail below.

\begin{table}[bt]
\begin{center}
\scalebox{0.82}{
\begin{tabular}{llrrrr} \hline
    Model&Train&Upward&Downward&Non&All\\ \hline \hline
    \multicolumn{2}{c}{Majority}&65.5&63.3&99.3&50.4\\ \hline
    BiMPM&SNLI&53.5&\textbf{57.6}&27.4&\textbf{54.6}\\
    ESIM&SNLI&71.1&45.2&41.8&53.8\\
    DeComp&SNLI&66.1&42.1&\textbf{64.4}&51.4\\
    KIM&SNLI&78.8&30.3&53.1&48.0\\
    BERT&SNLI&50.1&46.8&7.5&45.8\\ 
    BERT&MNLI&\textbf{82.7}&22.8&52.7&44.7\\ \hline
\end{tabular}
}
\caption{\label{tab:perform} Accuracies (\%) for different models and training datasets.}
\end{center}
\end{table}

\subsection{Data augmentation for analysis}
To explore whether the performance of models on monotonicity reasoning depends on the training set or the model themselves,
we conducted further analysis performed by data augmentation with the automatically generated monotonicity dataset HELP~\cite{yanaka2019}.
HELP contains 36K monotonicity inference examples (7,784 upward examples, 21,192 downward examples, and 1,105 non-monotone examples).
The size of the HELP word vocabulary is 15K, and the overlap ratio of vocabulary between HELP and MED is 15.2\%.

We trained BERT on MultiNLI only and on MultiNLI augmented with HELP,
and compared their performance.
Following \final{\citet{poliak-EtAl:2018:S18-2}}, we also checked the performance of a hypothesis-only model trained with each training set to test whether our test set contains undesired biases.

\begin{table}[bt]
\begin{center}
\scalebox{0.80}{
\begin{tabular}{l|rrrr} \hline
    Training set&Upward&Downward&Non&All\\ \hline \hline
     MNLI&\textbf{82.7}&22.8&52.7&44.7\\
     MNLI--Hyp&34.3&18.3&31.5&24.4\\
     MNLI+HELP&76.0&\textbf{70.3}&\textbf{59.9}&\textbf{71.6}\\ 
     MNLI+HELP--Hyp&61.3&30.5&34.9&41.1 \\

\hline
\end{tabular}
}
\caption{\label{tab:eval1} Evaluation results on types of monotonicity reasoning. --Hyp: Hypothesis-only model.}
\end{center}
\end{table}

\begin{figure*}
  \includegraphics[width=16cm, bb=0.000000 0.000000 8876.000000 2498.000000]{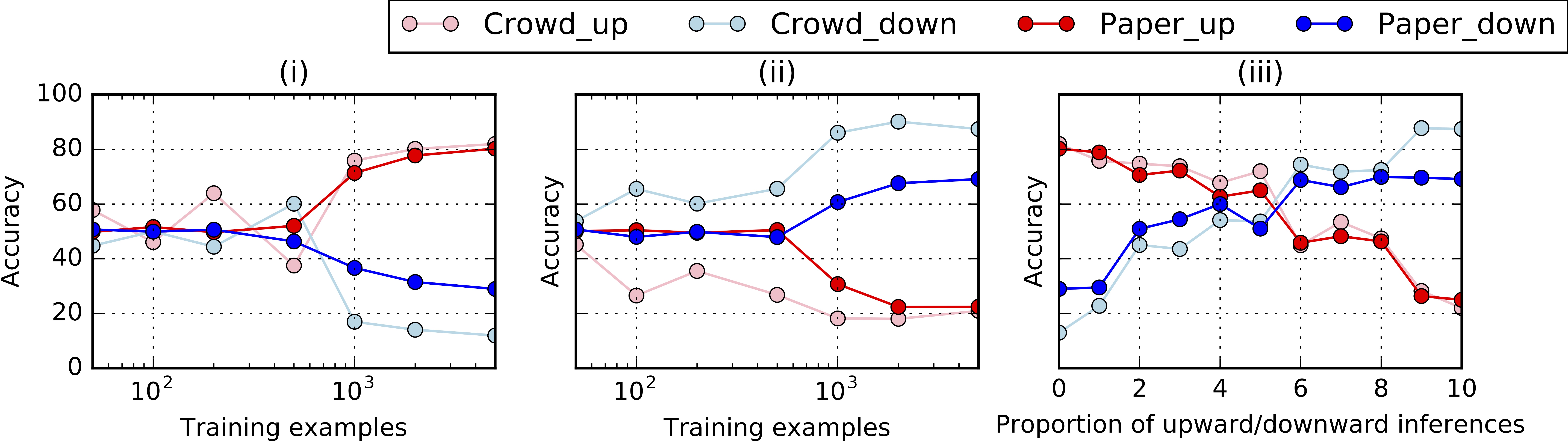}
  \caption{\label{tab:fig1} Accuracy throughout training BERT (i) with only upward examples and (ii) with only downward examples. We checked the accuracy at sizes [50, 100, 200, 500, 1000, 2000, 5000] for each direction. (iii) Performance on different ratios of upward/downward training sets. The total size of the training sets was 5,000 examples.}
\end{figure*}

\subsubsection{Effects of data augmentation}
Table~\ref{tab:eval1} shows that the performance of BERT with the hypothesis-only training set dropped around 10-40\% as compared with the one with the premise-hypothesis training set, even if we use the data augmentation technique.
This indicates that the MED test set does not allow models to predict from hypotheses alone.
Data augmentation by HELP improved the overall accuracy to 71.6\%, but there is still room for improvement.
In addition, while adding HELP increased the accuracy on downward inferences, it slightly decreased accuracy on upward inferences.
The size of downward examples in HELP is much larger than that of upward examples.
This might improve accuracy on downward inferences, but might decrease accuracy on upward inferences.

To investigate the relationship between accuracy on upward inferences and downward inferences, we checked the performance throughout training BERT with only upward and downward inference examples in HELP (Figure~\ref{tab:fig1} (i), (ii)).
These two figures show that, as the size of the upward training set increased, BERT performed better on upward inferences but worse on downward inferences, and vice versa.

Figure~\ref{tab:fig1} (iii) shows performance on a different ratio of upward and downward inference training sets. 
When downward inference examples constitute more than half of the training set, accuracies on upward and downward inferences were reversed.
As the ratio of downward inferences increased, BERT performed much worse on upward inferences.
This indicates that a training set in one direction (upward or downward entailing) of monotonicity might be harmful to models when learning the opposite direction of monotonicity.

Previous work using HELP~\cite{yanaka2019} reported that the BERT trained with MultiNLI and HELP containing both upward and downward inferences improved accuracy on both directions of monotonicity.
\final{MultiNLI rarely comes from downward inferences (see Section~\ref{sec:discuss}), and its size is large enough to be immune to the side-effects of downward inference examples in HELP.}
This indicates that MultiNLI might act as a \textit{buffer} against side-effects of the monotonicity-driven data augmentation technique.



\subsubsection{Linguistics-oriented versus human-oriented}
Table~\ref{tab:eval4} shows the evaluation results by genre.
This result shows that inference problems collected from linguistics publications are more challenging than crowdsourced inference problems, even if we add HELP to training sets.
As shown in Figure~\ref{tab:fig1}, the change in performance on problems from linguistics publications is milder than that on problems from crowdsourcing.
This result also indicates the difficulty of problems from linguistics publications.
Regarding non-monotone problems collected via crowdsourcing, there are very few non-monotone problems, so accuracy is 100\%.
Adding non-monotone problems to our test set is left for future work.

\begin{table}[bt]
\begin{center}
\scalebox{0.75}{
\begin{tabular}{lr|rrr} \hline
Genre&&$-$HELP&$+$HELP&$\triangle$\\ \hline \hline
\multirow{4}{*}{Crowd}&Up&87.1&83.6&$-$3.5\\
&Down&21.2&70.3&$+$49.1\\
&Non&100.0&100.0&$\pm$0.0\\ \cline{2-5}
&All&40.9&74.3&$+$33.4\\ \hline
\multirow{4}{*}{Paper}&Up&74.5&60.8&$-$13.7\\
&Down&33.8&69.5&$+$35.7\\
&Non&52.4&59.7&$+$7.3\\ \cline{2-5}
&All&56.6&63.3&$+$6.7\\ \hline
\end{tabular}
}
\caption{\label{tab:eval4} Evaluation results by genre. Paper: problems collected from linguistics publications, Crowd: problems via crowdsourcing.}
\end{center}
\end{table}

\subsubsection{Linguistic phenomena}
Table~\ref{tab:eval3new} shows the evaluation results by type of linguistic phenomenon.
While accuracy on problems involving NPIs and conditionals was improved on both upward and downward inferences, accuracy on problems involving conjunction and disjunction was improved on only one direction.
In addition, it is interesting to see that the change in accuracy on conjunction was opposite to that on disjunction.
Downward inference examples involving disjunction are similar to upward inference ones; that is, inferences from a sentence to a shorter sentence are valid (e.g., \textit{Not many campers have had a sunburn or caught a cold} $\Rightarrow$
\textit{Not many campers have caught a cold}).
Thus, these results were also caused by addition of downward inference examples.
Also, accuracy on problems annotated with \textit{reverse} tags was apparently better without HELP because all examples are upward inferences embedded in a downward environment twice.

Table~\ref{tab:eval3new} also shows that accuracy on conditionals was better on upward inferences than that on downward inferences.
This indicates that BERT might fail to capture the monotonicity property that conditionals create a downward entailing context in their scope while they create an upward entailing context out of their scope.

Regarding lexical knowledge, the data augmentation technique improved the performance much better on downward inferences which do not require lexical knowledge.
However, among the 394 problems for which all models provided wrong answers, 244 problems are non-lexical inference problems.
This indicates that some non-lexical inference problems are more difficult than lexical inference problems, though accuracy on non-lexical inference problems was better than that on lexical inference problems.



\begin{table}[bt]
\begin{center}
\scalebox{0.75}{
\begin{tabular}{llr|rrr} \hline
Tag&&&$-$HELP&$+$HELP&$\triangle$\\ \hline \hline
\multirow{7}{*}{Up}&Lexical&(743)&81.0&70.8&$-$10.2\\
&non-Lexical&(1,077)&84.1&79.6&$-$4.5\\
&NPIs&(64)&20.3&35.9&$+$15.6\\
&Conditionals&(29)&51.7&62.1&$+$9.4\\
&Conjunction&(175)&94.3&88.0&$-$6.3\\
&Disjunction&(96)&4.2&32.3&$+$28.1\\
&Reverse&(240)&74.2&28.7&$-$45.5\\ \hline
\multirow{6}{*}{Down}&Lexical&(477)&46.1&64.6&$+$18.5\\
&non-Lexical&(2,793)&18.8&71.2&$+$52.4\\
&NPIs&(266)&44.0&60.2&$+$16.2\\
&Conditionals&(120)&15.8&20.0&$+$4.2\\
&Conjunction&(106)&24.5&40.6&$+$16.1\\
&Disjunction&(138)&80.4&40.6&$-$39.8\\ \hline
\multirow{4}{*}{Non}&Lexical&(182)&58.2&64.3&$+$6.1\\
&non-Lexical&(110)&43.6&52.7&$+$9.1\\
&NPIs&(8)&0.0&0.0&$\pm$0.0\\
&Disjunction&(20)&10.0&15.0&$+$5.0\\
\hline
\end{tabular}
}
\caption{\label{tab:eval3new} Evaluation results by linguistic phenomenon type. (non-)Lexical: problems that (do not) require lexical relations. Numbers in parentheses are numbers of problems.}
\end{center}
\end{table}






\subsection{Discussion}
\label{sec:discuss}

One of our findings is that there is a type of downward inferences to which every model fails to provide correct answers. One such example is concerned with the contrast between \textit{few} and \textit{a few}.
%
Among 394 problems for which all models provided wrong answers, 148 downward inference problems were problems involving the downward monotonicity operator \textit{few} such as in the following example:

{\footnotesize
\begin{exe}
\ex \label{ex:9}
 \begin{xlist}
    \exi{$P$:} \textit{Few of the books had typical or marginal readers}
    \exi{$H$:} \textit{Few of the books had some typical readers}
 \end{xlist}
\end{exe}
}
We transformed these downward inference problems to upward inference problems in two ways: (i) by replacing the downward operator \textit{few} with the upward operator \textit{a few}, and (ii) by removing the downward operator \textit{few}.
We tested BERT using these transformed test sets.
The results showed that BERT predicted the same answers for the transformed test sets.
This suggests that BERT does not understand the difference between the downward operator \textit{few} and the upward operator \textit{a few}.

The results of crowdsourcing tasks in Section 3.1.3 showed that some downward inferences can naturally be performed in human reasoning.
However, we also found that the MultiNLI training set~\cite{DBLP:journals/corr/WilliamsNB17}, which is one of the dataset created from naturally-occurring texts, contains only 77 downward inference problems,
including the following one.\footnote{
The MultiNLI training set
has 1,700 inference problems where the downward entailing operators \textit{no} and \textit{never} occur in both the premise and the hypothesis, but most of them
are not an instance of downward inferences.}
%

{\footnotesize
\begin{exe}
\ex \label{ex:10}
 \begin{xlist}
    \exi{$P$:} \textit{No racin' on the Range}
    \exi{$H$:} \textit{No horse racing is allowed on the Range}
 \end{xlist}
\end{exe}
}

%
%

%

\noindent
One possible reason why there are few downward inferences is that certain pragmatic factors can block people to draw a downward inference.
For instance, in the case of the inference problem in  (\ref{ex:11}), unless the added disjunct in $H$, i.e., \textit{a small cat with green eyes}, is salient in the context, it would be difficult to draw the conclusion $H$ from the premise $P$.

{\footnotesize
\begin{exe}
\ex \label{ex:11}
 \begin{xlist}
    \exi{$P$:} \textit{I saw a dog}
    \exi{$H$:} \textit{I saw a dog or a small cat with green eyes}
 \end{xlist}
\end{exe}
}

\noindent
Such pragmatic factors would be one of the reasons why it is difficult to obtain downward inferences in naturally occurring texts.

\section{Conclusion}
\label{sec:conc}
We introduced a large monotonicity entailment dataset, called MED.
To illustrate the usefulness of MED, we tested state-of-the-art NLI models, and found that performance on the new test set was substantially worse for all state-of-the-art NLI models.
In addition, the accuracy on downward inferences was inversely proportional to the one on upward inferences.

An experiment with the data augmentation technique showed that accuracy on upward and downward inferences depends on the proportion of upward and downward inferences in the training set.
This indicates that current neural models might have limitations on their generalization ability in monotonicity reasoning.
We hope that the MED will be valuable for future research on more advanced models that are capable of monotonicity reasoning in a proper way.

\section*{Acknowledgement}
This work was partially supported by JST AIP- PRISM Grant Number JPMJCR18Y1, Japan, and JSPS KAKENHI Grant Number JP18H03284, Japan. 
We thank our three anonymous reviewers for helpful suggestions. We are also grateful to Koki Washio, Masashi Yoshikawa, and Thomas McLachlan for helpful discussion.

\bibliographystyle{acl_natbib}
\bibliography{acl2019}

\end{document}